\documentclass[10pt,conference]{IEEEtran}
\usepackage{epsfig,rotating,setspace,latexsym,amsmath,epsf,amssymb,amsfonts,bm,theorem,cite,authblk, bbm,color, hyperref, multirow, caption, subcaption, algorithm, algpseudocode}

\DeclareMathOperator*{\argmax}{arg\,max}

\IEEEoverridecommandlockouts
\allowdisplaybreaks

\title{Low-Latency Task-Oriented Communications with Multi-Round, Multi-Task Deep Learning}
\begin{document}
\author[1]{Yalin E. Sagduyu}
\author[1]{Tugba Erpek}
\author[2]{Aylin Yener}
\author[3]{Sennur Ulukus}

\affil[1]{\normalsize  Nexcepta, Gaithersburg, MD, USA}

\affil[2]{\normalsize  The Ohio State University, Columbus, OH, USA}

\affil[3]{\normalsize University of Maryland, College Park, MD, USA}
\maketitle

\begin{abstract}
In this paper, we address task-oriented (or goal-oriented) communications where an encoder at the transmitter learns compressed latent representations of data, which are then transmitted over a wireless channel. At the receiver, a decoder performs a machine learning task, specifically for classifying the received signals. The deep neural networks corresponding to the encoder-decoder pair are jointly trained, taking both channel and data characteristics into account. Our objective is to achieve high accuracy in completing the underlying task while minimizing the number of channel uses determined by the encoder's output size. To this end, we propose a multi-round, multi-task learning (MRMTL) approach for the dynamic update of channel uses in multi-round transmissions. The transmitter incrementally sends an increasing number of encoded samples over the channel based on the feedback from the receiver, and the receiver utilizes the signals from a previous round to enhance the task performance, rather than only considering the latest transmission. This approach employs multi-task learning to jointly optimize accuracy across varying number of channel uses, treating each configuration as a distinct task. By evaluating the confidence of the receiver in task decisions, MRMTL decides on whether to allocate additional channel uses in multiple rounds. We characterize both the accuracy and the delay (total number of channel uses) of MRMTL, demonstrating that it achieves the accuracy close to that of conventional methods requiring large numbers of channel uses, but with reduced delay by incorporating signals from a prior round. We consider the CIFAR-10 dataset, convolutional neural network architectures, and AWGN and Rayleigh channel models for performance evaluation. We show that MRMTL significantly improves the efficiency of task-oriented communications, balancing accuracy and latency effectively. 
\end{abstract}


\section{Introduction}
In the rapidly advancing field of NextG communication systems, there is an increasing focus on task-oriented (or goal-oriented) communications. This approach is gaining prominence as it addresses the specific needs of various applications by ensuring that the transmission process is aligned with the ultimate objective of the task at hand \cite{GunduzBeyondBits2023, Calvanese6G2023, XieTOC2022, FarshbafanTOC2022, shiTOC2023, LiuTOC2024, GetuTOC2024, shao2021learning, zhou2024goal}. Unlike traditional communication paradigms that focus on delivering raw data, task-oriented communications (TOC) aims to transmit only the information necessary to accomplish a specific task. Deep learning plays a crucial role in optimizing the encoding and decoding processes for TOC, allowing for efficient and effective transmission of information that directly contributes to the task’s success. By leveraging deep learning-driven TOC, NextG communication systems can achieve significant improvements in both performance and resource utilization \cite{sagduyuTOCWirCom2023, sagduyuMultiRXTOCFNWF2023, sagduyuSemComNetMag2024}, making them well-suited for the demands of modern applications such as the Internet of Things (IoT), augmented reality/virtual reality (AR/VR), and vehicle-to-everything (V2X) network systems.

In IoT networks, sensors generate vast amounts of data that need to be processed and analyzed to make real-time decisions, such as in smart cities and industrial automation. TOC can significantly reduce the communication overhead by transmitting only the essential information required for decision-making, rather than the raw sensor data. Similarly, in AR/VR applications, low latency and high accuracy are critical to delivering immersive experiences. TOC can help achieve this by optimizing the transmission of visual and sensory data to meet the application's specific needs. In V2X systems, vehicles need to communicate with each other and with infrastructure to ensure safe and efficient transportation. TOC can enhance these interactions by focusing on the transmission of critical information, such as collision warnings and traffic updates, thereby improving response times and reducing network congestion.

One of the primary challenges in TOC is balancing task accuracy and latency objectives and requirements. To that end, the age of task information for TOC was studied in \cite{sagduyuAoIInfocom2023}. Achieving high accuracy often requires transmitting a large amount of data, which can lead to increased delay (measured by the number of channel uses) and higher bandwidth usage. Conversely, minimizing delay and bandwidth usage can compromise accuracy. This accuracy-delay tradeoff is a significant hurdle that needs to be addressed to realize the full potential of TOC.

We propose a novel multi-round, multi-task learning (MRMTL) approach to address this challenge by dynamically updating the number of channel uses in iterative transmissions of TOC. MRMTL involves an encoder at the transmitter that learns compressed latent representations of input data (e.g., images), which are transmitted over a wireless channel. At the receiver, a decoder performs a machine learning task, specifically classifying the received signals. MRMTL is different from the autoencoder-based communications, where the typical setting has the source-coded data symbols as the input and the reconstruction of those symbols as the output \cite{erpekDLComm2020}. On the other hand, MRMTL starts with (raw) input data and performs a machine learning task such as classification. The deep neural networks (DNNs) corresponding to the encoder-decoder pair are jointly trained, considering both channel and data characteristics to achieve high task accuracy with minimal channel uses.

MRMTL introduces a dynamic update mechanism where the transmitter incrementally sends an increasing number of encoded samples over the channel. The receiver utilizes signals from a previous round to enhance task performance, rather than relying solely on the latest transmissions. The multi-round process utilizes multi-task learning that jointly optimizes accuracy across multiple rounds, with each configuration treated as a distinct task performed with a different number of channel uses. When the receiver's confidence in the task decisions is low, it can then allocate additional channel uses to improve task accuracy. We demonstrate that MRMTL achieves the accuracy of conventional methods with single-round, single-task learning (SRSTL) for TOC that requires a large number of channel uses, but with significantly reduced delay by incorporating signals from a prior round.

To evaluate MRMTL, we consider the CIFAR-10 dataset as the input, convolutional neural network (CNN) architectures as the DNNs, and AWGN and Rayleigh channel models. Our results show that MRMTL significantly improves the efficiency of TOC, effectively balancing accuracy and delay. This study represents a significant step forward in the development of efficient and effective TOC systems for NextG networks. The MRMTL approach can be extended to incorporate semantic communications \cite{sagduyuSemComMag2023} and integrated sensing and communications \cite{sagduyuJSaCDySpan2024, sagduyuJSaCarXiv2023} by adding tasks of reconstruction and sensing, respectively.  

The remainder of the paper is organized as follows. Section \ref{sec:singletask} describes SRSTL for TOC. Section \ref{sec:multitask} presents MRMTL for TOC. Section \ref{sec:threshold} introduces the MRMTL's process of dynamic initiation of multiple rounds in TOC.
\begin{figure}[t!]
	\centering
	\includegraphics[width=\columnwidth]{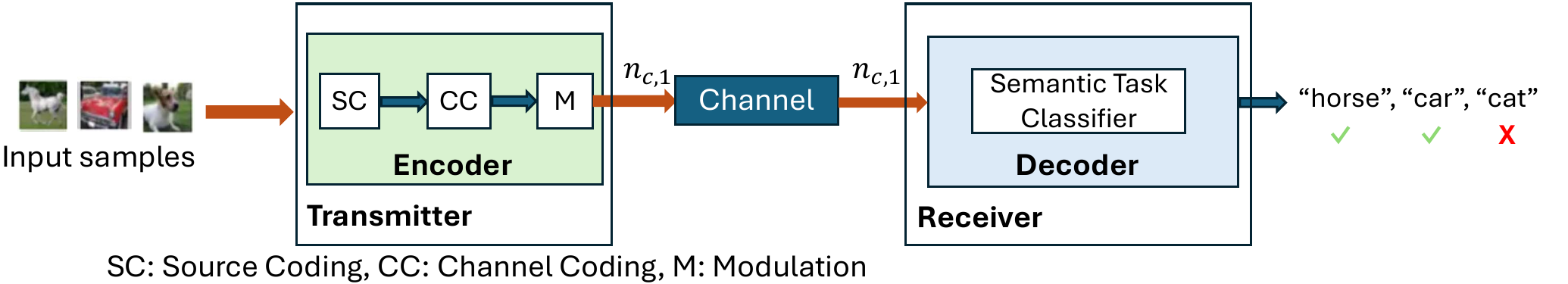}
	\caption{System model of SRSTL for TOC.}
	\label{fig:singlesysarch}
\end{figure}

\section{Single-Round, Single-Task Learning} \label{sec:singletask}
The system model of SRSTL for TOC is shown in Figure \ref{fig:singlesysarch}. 
The transmitter has input data samples (e.g., images) $\boldsymbol{x}$. The DNN-based Encoder $E_1$ at the transmitter performs the combined roles of source coding, channel coding and modulation, and outputs the modulated symbols,  $E_1(\boldsymbol{x})$, with size $n_{c,1}$.  The transmitted symbols go through the channel effects (over $n_{c,1}$ channel uses) and become the input to a Semantic Task Classifier at the receiver, as $ \boldsymbol{r} = \boldsymbol{h} \: E_1 (\boldsymbol{x}) + \boldsymbol{n}$, where $\boldsymbol{h}$ is the channel gain and $\boldsymbol{n}$ is the noise. The Decoder, $D_1$, corresponding to the Semantic Task Classifier is also a DNN whose output is the predicted labels, $\boldsymbol{\hat{y}}_1 = D_1(\boldsymbol{r} )$. The task is accomplished successfully when $\boldsymbol{\hat{y}}_1 = \boldsymbol{y}^*$, namely the received signals are correctly classified as the true labels $\boldsymbol{y}^*$. The task accuracy increases with $n_{c,1}$ as demonstrated in \cite{sagduyuTOCWirCom2023, sagduyuMultiRXTOCFNWF2023, sagduyuSemComNetMag2024}. By considering SRSTL as the baseline, we also highlight this trend for SRSTL in this section. On the other hand, the increase in $n_{c,1}$ also increases the latency (in terms of number of channel uses) for task completion. To reduce latency, we introduce MRMTL for TOC in Sec.~\ref{sec:multitask}.

For performance evaluation, we consider the CIFAR-10 dataset that consists of 60,000 color images, each with a resolution of 32x32 pixels, divided into 10 distinct classes as the transmitter input. Each class represents a different category, specifically: Airplane, Automobile, Bird, Cat, Deer, Dog, Frog, Horse, Ship, and Truck. The dataset is split into 50,000 training images and 10,000 test images, with 6,000 images per class. The DNN architectures for the Encoder and Decoder are shown in Table \ref{tab:NNArch}. 

\begin{table}[ht]
\footnotesize
    \centering
    \caption{DNN architectures for TOC.}
    \label{tab:NNArch}
    \begin{subtable}[t]{0.5\textwidth}
    \centering
    \caption{Encoder DNN}
    \begin{tabular}{l||l}
    \hline
    Layers & Properties \\ \hline
       Conv2D & filters = 32, kernel size = (3,3), \\ & activation function = ReLU \\
       Conv2D & filters = 32, kernel size = (3,3), \\ & activation function = ReLU \\
       Max pooling & pool size = (2,2) \\
       Dropout & dropout rate = $0.25$ \\
       Conv2D & filters = 64, kernel size = (3,3), \\ & activation function = ReLU \\
       Conv2D & filters = 64, kernel size = (3,3), \\ & activation function = ReLU \\
       Max pooling & pool size = (2,2) \\
       Dropout & 0.25 \\
       Conv2D & filters = 128, kernel size = (3,3), \\ & activation function = ReLU \\
       Conv2D & filters = 128, kernel size = (3,3), \\ & activation function = ReLU \\
       Max pooling & pool size = (2,2) \\
       Dropout & 0.25 \\    
       Flatten & -- \\
       Dense & size = 512, activation function = ReLU\\
       Dropout & dropout rate = $0.25$ \\
       Dense & size = $n_{c,1}$ for Round 1 or $n_{c,2}$ for Round 2, \\ & activation function = linear \\
    \end{tabular}
    \vspace{0.25cm}
    \end{subtable}
    \begin{subtable}[t]{0.5\textwidth}
    \centering
    \caption{Decoder DNN}
    \begin{tabular}{l||l}
    \hline
    Layers & Properties \\ \hline
       Dense  & size = $n_{c,1}$ for Round 1 or $n_{c,1} + n_{c,2}$ for Round 2, \\ & activation function = ReLU \\
       Dropout & dropout rate = $0.1$ \\
       Dense & size = $n_{c}$, activation function = ReLU \\
       Dense & size = $10$, activation function  = SoftMax \\ \hline
    \end{tabular}
    \end{subtable} 
\end{table}

Task accuracy of TOC with SRSTL is shown in Table~\ref{tab:accuracy_single} under AWGN and Rayleigh channels.  Note that we keep the distribution the same but draw a different sample from the distribution in each training and test instance for AWGN and Rayleigh channels. Task accuracy of SRSTL for TOC increases when $n_{c,1}$ increases from $n_c$ to $2n_c$, or when channel is changed from Rayleigh to AWGN. Since optimization is harder for TOC training under Rayleigh channel, we set $n_c$ as $5$ and $16$ for AWGN and Rayleigh channels, respectively.

\begin{table}[ht]
\footnotesize
    \centering
    \caption{Average task accuracy of TOC with SRSTL.}    
    \label{tab:accuracy_single}
    \begin{tabular}{c||c|c}
    Number of channel uses & AWGN channel & Rayleigh channel \\
    \hline
    $n_{c,1}  = n_c$ & 0.7335 & 0.6359    \\   
    $n_{c,1}  = 2 n_c $ & 0.8166 & 0.7492 \\ \hline 
    \end{tabular}
    \vspace{0.5cm}
\end{table}

\section{Multi-Round, Multi-Task Learning} \label{sec:multitask}

\begin{figure}[t!]
	\centering
	\includegraphics[width=\columnwidth]{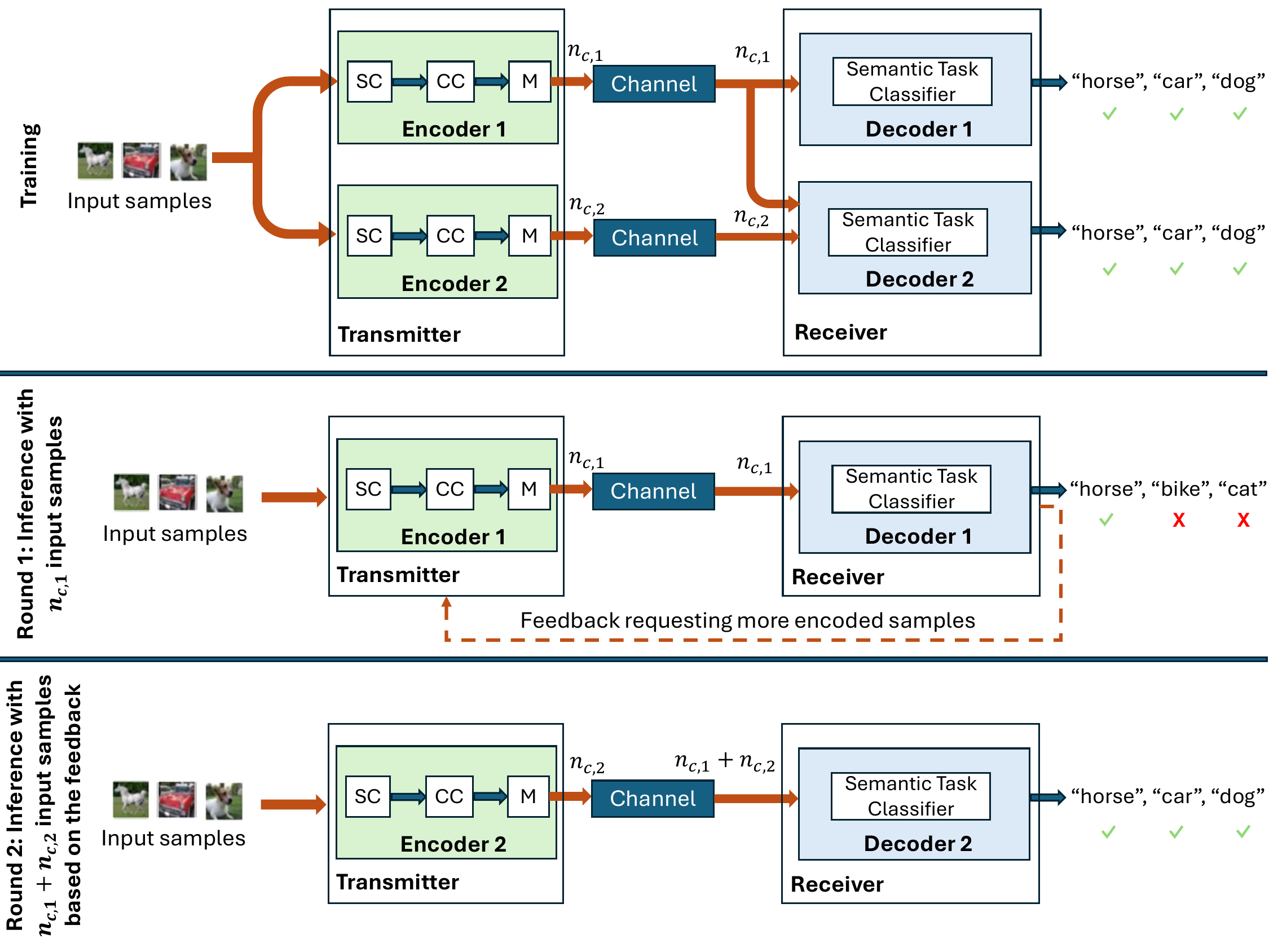}
	\caption{System model of MRMTL for TOC.}
	\label{fig:multisysarch}
\end{figure}

The system model of MRMTL for TOC is shown in Figure \ref{fig:multisysarch}. Multi-task learning is used to train Encoders $E_1$ and $E_2$, and Decoders $D_1$ and $D_2$ jointly. The loss function for the jointly trained system is defined as $l=w \: l_1+(1-w) \: l_2$, where $l_1$ is the loss from Decoder 1 in Round 1, $l_2$ is the loss from Decoder 2 in Round 2, and $w$ is the weight of loss $l_1$ during training ($w$ is set to $0.5$ for training in numerical results).

After training, in a latency-constrained system, we first start with transmitting a low number of samples as the transmitter output in Round 1. If the Semantic Task Classifier's accuracy is insufficient, then the receiver requests more encoded samples from the transmitter to improve the classification accuracy in Round 2. The earlier transmitted samples are still re-used to keep the latency low when the receiver needs more samples from the transmitter to improve the accuracy. To achieve this, the system is designed to operate in multiple rounds with a built-in feedback mechanism. 

Similar to SRSTL, the transmitter has input data samples $\boldsymbol{x}$. In Round 1, the transmitter encodes $\boldsymbol{x}$ with $E_1$ and transmits the encoded samples over $n_{c,1}$ channel uses. The received signals are $ \boldsymbol{r}_1 = \boldsymbol{h}_1 \: E_1 (\boldsymbol{x}) + \boldsymbol{n}_1$, where $\boldsymbol{h}_1$ is the channel gain and $\boldsymbol{n}_1$ is the noise in Round 1. The receiver classifies $\boldsymbol{r}_1$ with Decoder $D_1$ and predicts labels, $\boldsymbol{\hat{y}}_1 = D_1(\boldsymbol{r}_1 )$. In Round 2, the transmitter encodes $\boldsymbol{x}$ with $E_2$ and transmits the encoded samples over $n_{c,2}$ channel uses. The received signals are $ \boldsymbol{r}_2 = \boldsymbol{h}_2 \: E_2 (\boldsymbol{x}) + \boldsymbol{n}_2$, where $\boldsymbol{h}_2$ is the channel gain and $\boldsymbol{n}_2$ is the noise in Round 2. In Round 2, the receiver decodes the concatenation of signals received in both Rounds 1 and 2, $[\boldsymbol{r}_1, \boldsymbol{r}_2]$, with Decoder, $D_2$, and predicts labels, $\boldsymbol{\hat{y}}_2 = D_2([\boldsymbol{r}_1, \boldsymbol{r}_2])$.

The DNN architectures of the encoder-decoder pair for multi-task learning are similar to the architectures shown in Table \ref{tab:NNArch}. Encoder 1 and Encoder 2 provide $n_{c,1}$ and $n_{c,2}$ samples, respectively, as the output. Then the dense layer size is $n_{c,1}$ for Decoder 1 and $n_{c,1}+n_{c,2}$ for Decoder 2 at the receiver. We set $n_{c,1} = n_{c,2}$ for performance evaluation. 

\begin{figure}[ht]
\vspace{-1cm}
	\centering
        \includegraphics[width=0.925\columnwidth]{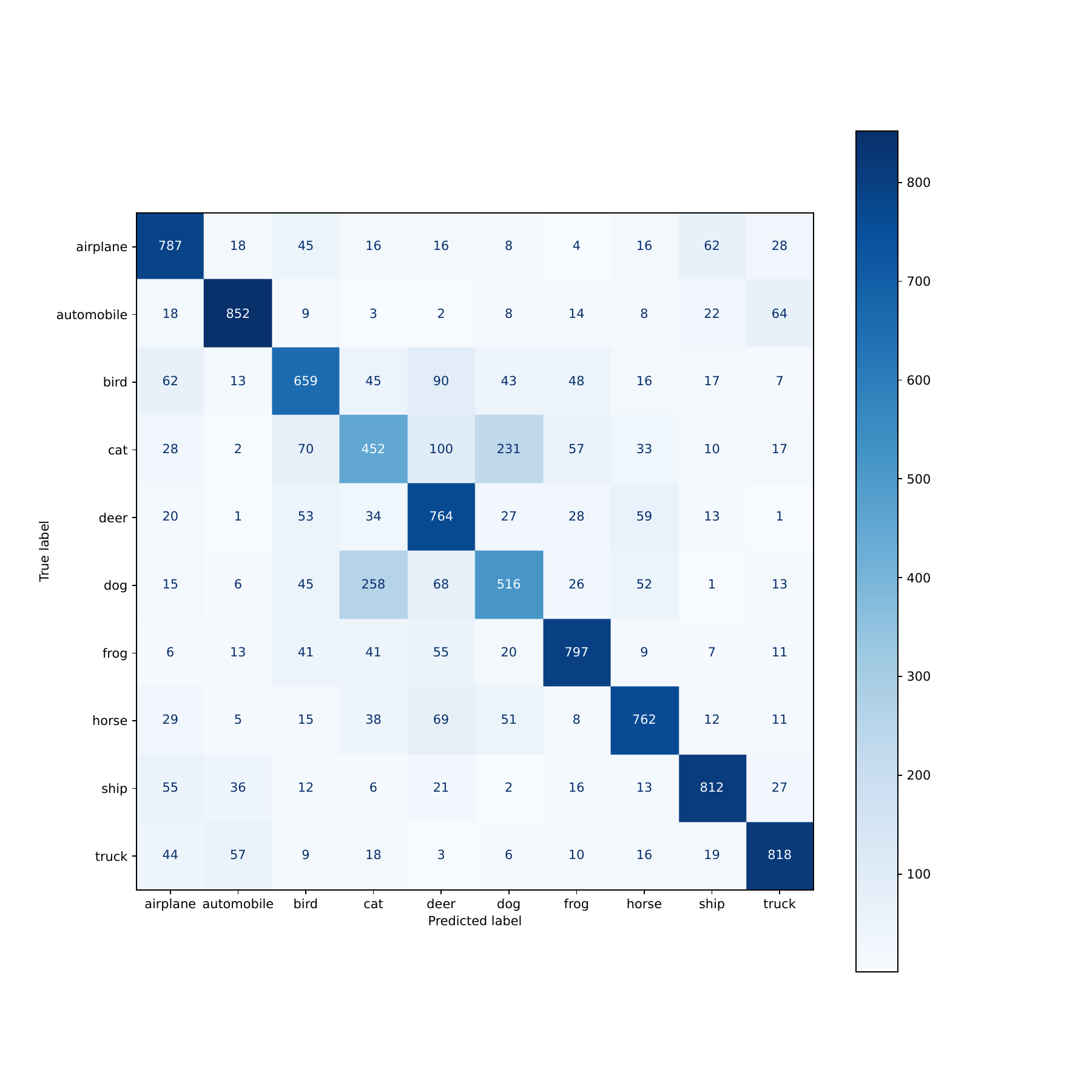}
     \vspace{-1cm}
    \caption{Confusion matrix for the first round of multi-task learning under AWGN channel.}
    \label{fig:CM1A}
\end{figure}

\begin{figure}[ht]
\vspace{-0.5cm}
	\centering
    \includegraphics[width=0.925\columnwidth]{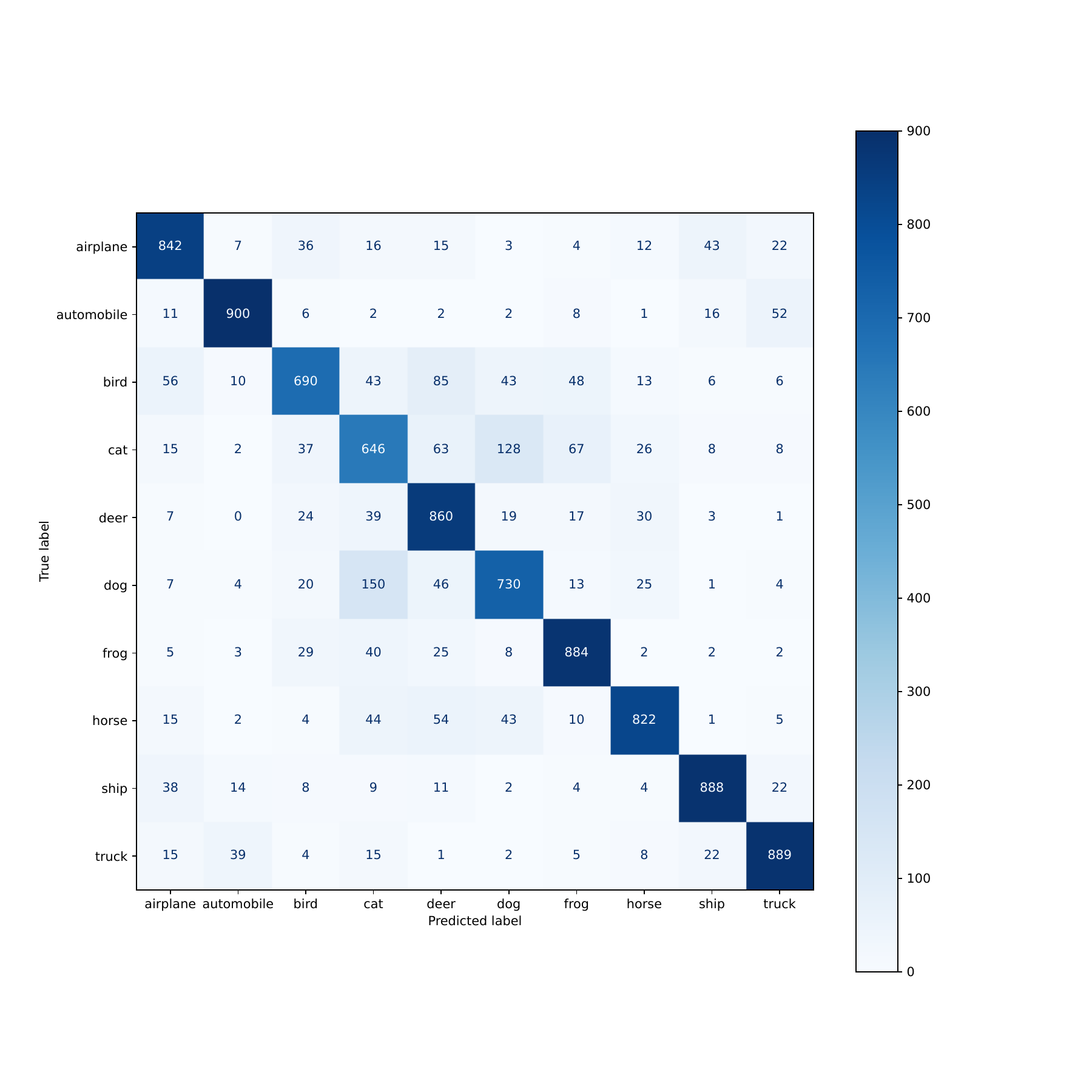}
     \vspace{-1cm}
    \caption{Confusion matrix for the second round of multi-task learning under AWGN channel.}
	\label{fig:CM2A}
\end{figure}

\begin{figure}[ht]
\vspace{-0.5cm}
	\centering
	\includegraphics[width=0.925\columnwidth]{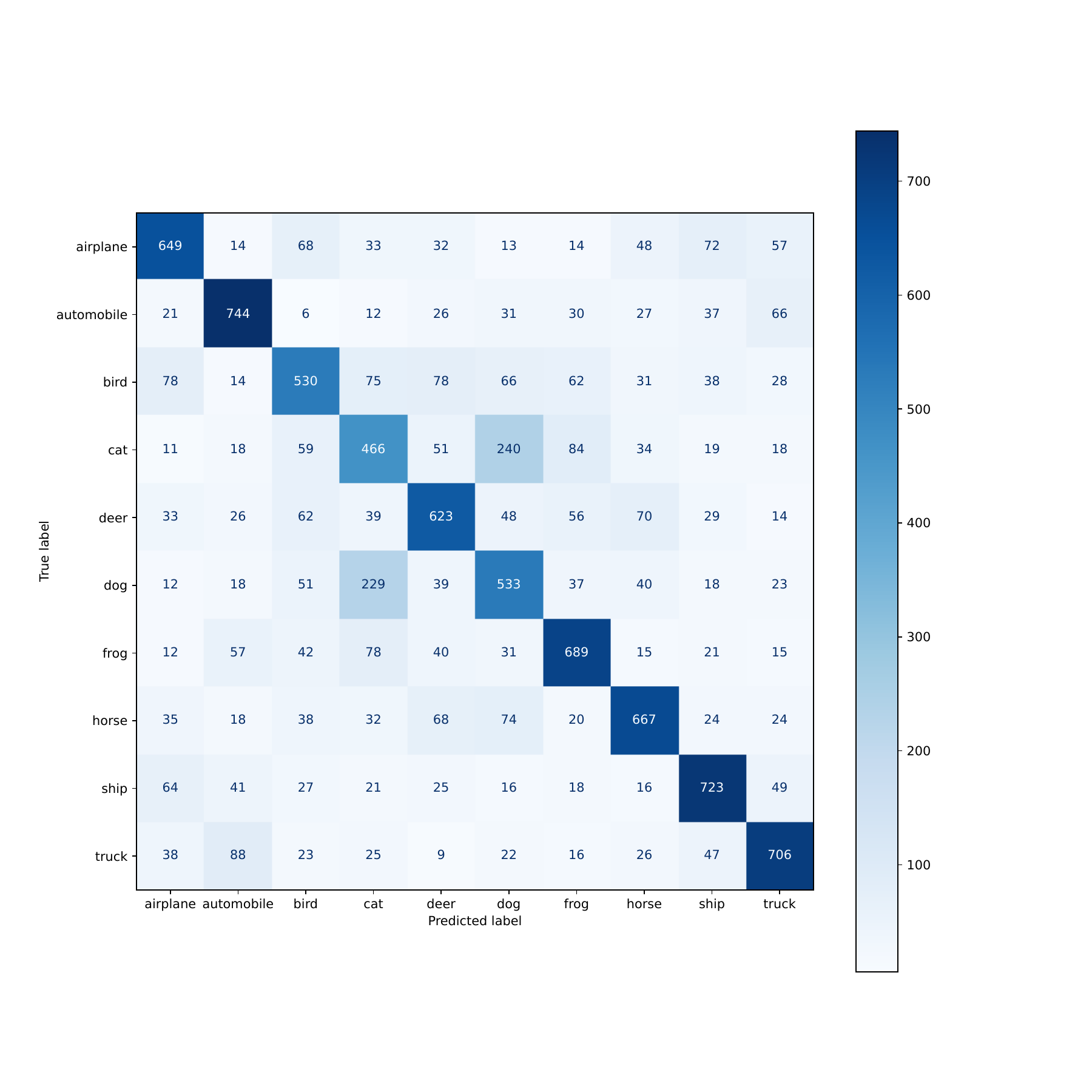}
     \vspace{-1cm}
    \caption{Confusion matrix for the first round of multi-task learning under Rayleigh channel.}
	\label{fig:CM1R}
\end{figure}

\begin{figure}[ht]
\vspace{-0.5cm}
	\centering
	\includegraphics[width=0.925\columnwidth]{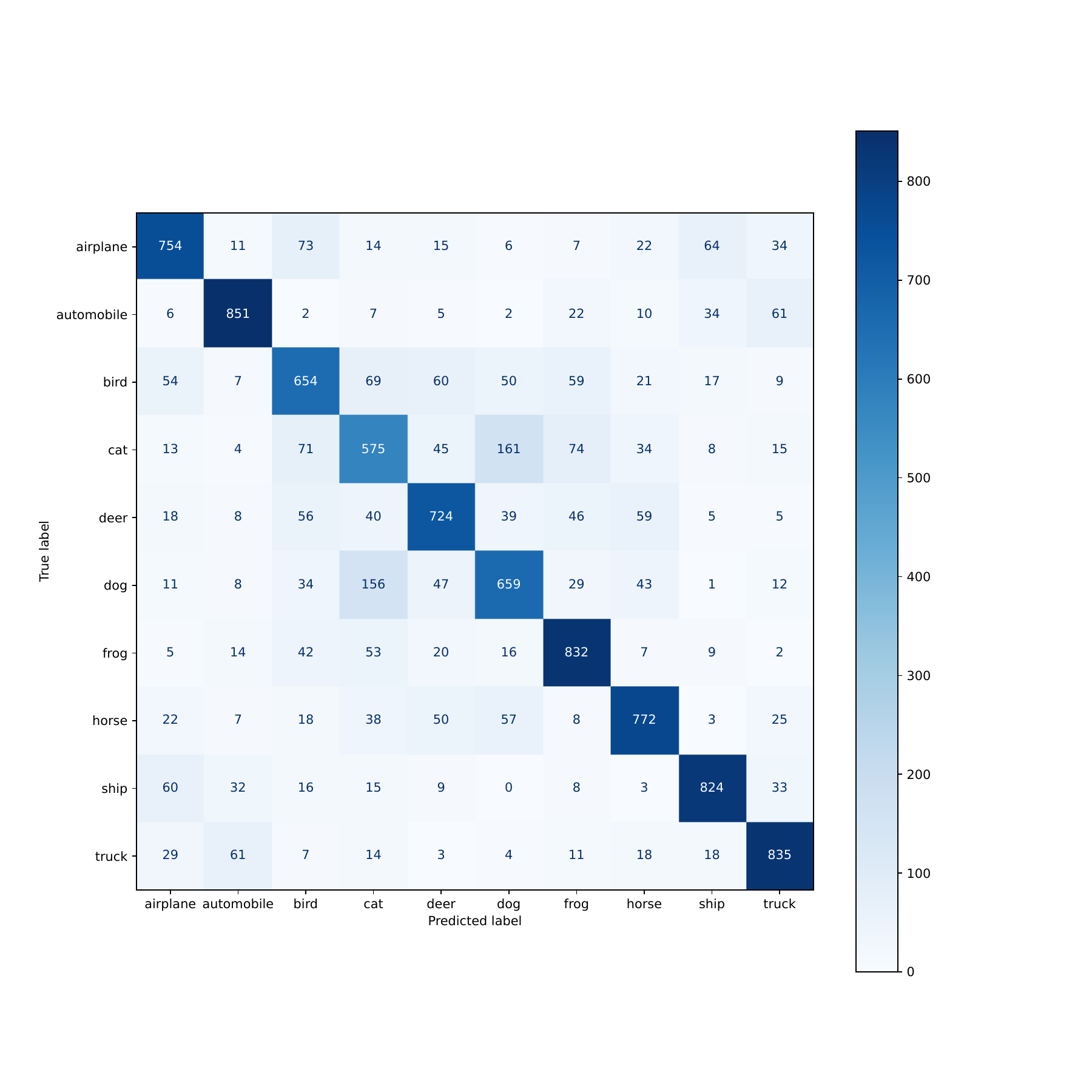}
     \vspace{-1cm}
    \caption{Confusion matrix for the second round of multi-task learning under Rayleigh channel.}
	\label{fig:CM2R}
\end{figure}

Task accuracy of TOC with MRMTL is shown in Table~\ref{tab:accuracy_multi} under AWGN and Rayleigh channels.   Task accuracy of MRMTL for TOC is higher for Round 2 (compared to Round 1) and under the AWGN channel (compared to Rayleigh channel).
Comparing results from Tables~\ref{tab:accuracy_single} and \ref{tab:accuracy_multi}, each round of MRMTL achieves task accuracy close to that of the corresponding SRSTL with the same number of channel uses. Confusion matrices for the first and second rounds of multi-task learning under the AWGN channel are shown in Figs.~\ref{fig:CM1A} and \ref{fig:CM2A}, respectively. Confusion matrices for the first and second rounds of multi-task learning under the Rayleigh channel are shown in Figs.~\ref{fig:CM1R} and \ref{fig:CM2R}, respectively. 

\begin{table}[ht]
\footnotesize
    \centering
    \caption{Average task accuracy of TOC with  MRMTL.}
    \label{tab:accuracy}
    \label{tab:accuracy_multi}
    \begin{tabular}{c||c|c}
    Round \# (number of channel uses) & AWGN channel & Rayleigh channel \\
    \hline
    Round 1 only ($n_{c,1} = n_c$) & 0.7320 & 0.6271     \\   
    Round 1 + Round 2 ($n_{c,1} + n_{c,2} = 2n_c$ ) & 0.8151 & 0.7433\\ \hline 
    \end{tabular}
    \vspace{0.5cm}
\end{table}

In Round 2 of MRMTL, the first $n_{c,1}$ samples are the output of Encoder 1 and the second $n_{c,2}$ samples are the output of Encoder 2 as opposed to a single encoder outputting $2n_c$ samples in single-task learning. MRMTL sustains similar task accuracy over $n_{c,1}+ n_{c,2}$ channel uses while decreasing latency compared to the case of SRSTL performed two times, the first one performed over $n_{c,1}$ channel uses and the second one performed over $n_{c,1}+ n_{c,2}$ channel uses (if the first one is not sufficient in terms of task accuracy). 

\section{Latency Reduction with Dynamic Multi-Round, Multi-Task Learning} \label{sec:threshold}
Adding Round 2 in MRMTL improves the task accuracy compared to the use of Round 1 only as in SRSTL. However, it also increases latency from $n_{c,1}$ to $n_{c,1} + n_{c,2}$. Next, we present a dynamic scheme that adaptively selects when to initiate Round 2 (as shown in Algorithm \ref{alg:cap}). Let $\boldsymbol{y}_{k,j,m}$ denote the $k$th entry of decoder $D_m$'s final layer output in Round $m=1,2$ of TOC for any input sample $\boldsymbol{x}_j$. To decide on a specific class, we select the maximum of decoder's final layer output and map it to the corresponding class type. Specifically, $\boldsymbol{\hat{y}}_{j,m} = \argmax_k \boldsymbol{y}_{k,j,m} $ denotes the output label of decoder $D_m$ of Round $m = 1,2$ in TOC for any input sample $\boldsymbol{x}_j$. 

\begin{figure}[ht]
     \vspace{-0.5cm}
	\centering
	\includegraphics[width=0.925\columnwidth]{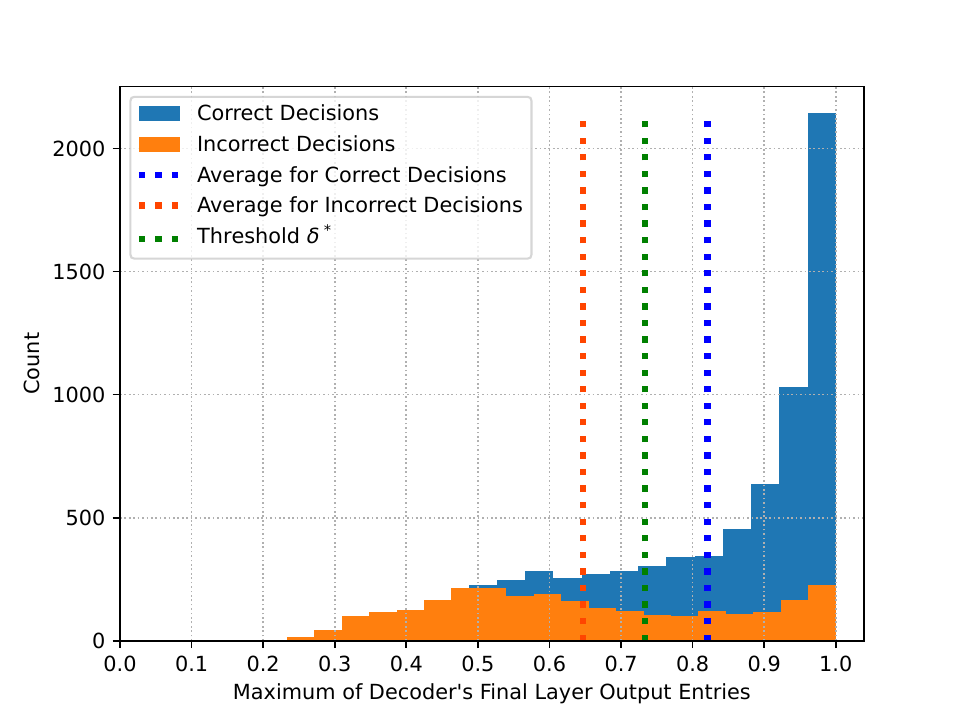}
    \caption{Histogram of the maximum of decoder final layer output vectors under AWGN channel.}
	\label{fig:histA}
\end{figure}

\begin{figure}[ht]
	\centering
      \vspace{-0.5cm}
	\includegraphics[width=0.925\columnwidth]{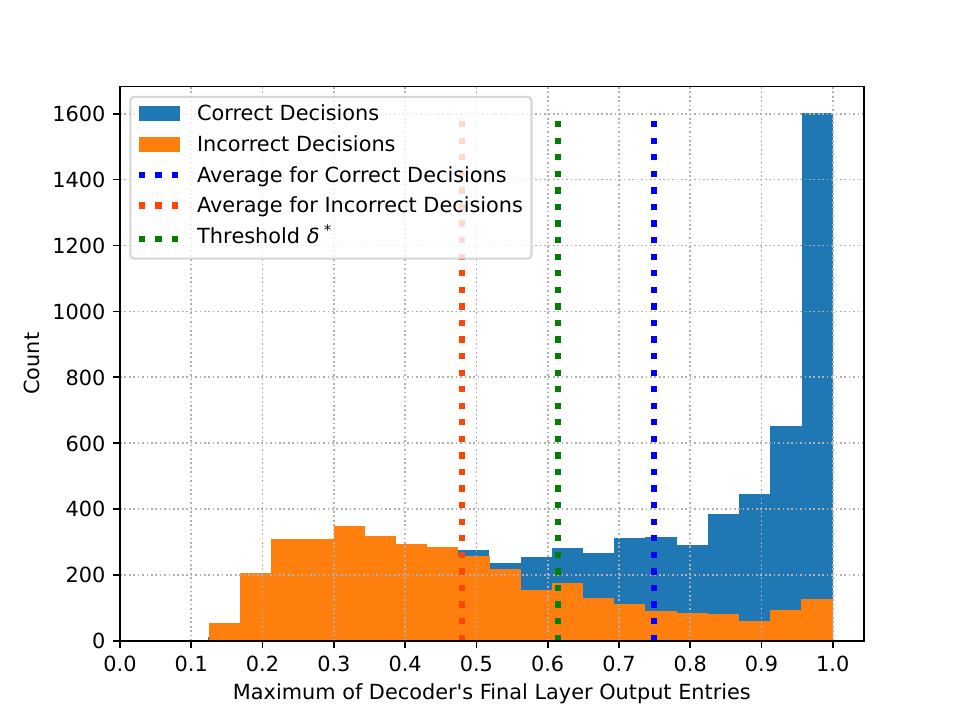}
    \caption{Histogram of the maximum of decoder final layer output vectors under Rayleigh channel.}
	\label{fig:histR}
\end{figure}

We define $\boldsymbol{y}_{j,m}^{\max} = \max_k \boldsymbol{y}_{k,j,m}$. Histograms of $\boldsymbol{y}_{1,j}^{\max}$ values under AWGN and Rayleigh channels are shown in Figs.~\ref{fig:histA} and \ref{fig:histR}, respectively. We denote $\boldsymbol{y}_1^{\max}$ as the random variable corresponding to the maximum of decoder's final layer output entries for a given input sample, $\boldsymbol{\hat{y}}_{1}$ as the label predicted in Round 1, and $\boldsymbol{y}^*$ as the respective true label. Overall, distribution of $\boldsymbol{y}_1^{\max}$ differs significantly under cases when correct and incorrect decisions are made in the first round (namely, when $\boldsymbol{\hat{y}}_{1} = \boldsymbol{y}^*$ and $\boldsymbol{\hat{y}}_{1} \neq \boldsymbol{y}^*$), showing different levels of confidence in decision-making. As $\boldsymbol{y}_1^{\max}$ increases, the accuracy tends to increase, as well. Therefore, at the end of Round 1, $\boldsymbol{y}_{j,1}^{\max}$ values closer to 1 indicate a more confident decision in predicting label $\boldsymbol{\hat{y}}_{j,1}$ for input sample $\boldsymbol{x}_{j}$. For any input sample $\boldsymbol{x}_{j}$, the receiver compares $\boldsymbol{y}_{j,1}^{\max}$ with a threshold $\delta$. If $\boldsymbol{y}_{j,1}^{\max} < \delta$, the receiver sends a feedback to the transmitter to initiate Round 2 transmissions. Otherwise, Round 2 is not initiated and $\boldsymbol{\hat{y}}_1 = D_1(\boldsymbol{r}_1)$ becomes the predicted label. If Round 2 is initiated, $\boldsymbol{\hat{y}}_2 = D_2([\boldsymbol{r}_1, \boldsymbol{r}_2])$ becomes the predicted label. This way, Round 2 is initiated only when the receiver is not confident on the classification in Round 1.   

\begin{algorithm}
\caption{MRMTL algorithm for TOC.}\label{alg:cap}
\begin{algorithmic}[1]
\State For a given test input sample, $\boldsymbol{x}_j$:
\State Round 1: Transmit encoder output $E_1[\boldsymbol{x}_j]$ of Task 1 with $n_{c,1}$ channel uses. \label{op0} 
    \If{$  \boldsymbol{y}_{j,1}^{\max}  < \delta$ }
        \State Initiate Round 2: Transmit encoder output $E_2[\boldsymbol{x}_j]$ of Task 2 with $n_{c,2}$ channel uses.
        \State Classify signals  $[\boldsymbol{r}_{j,q}, \boldsymbol{r}_{j,2}]$, received over $n_{c,1} + n_{c,2}$ channel uses of Round 1 and Round 2 such that $\boldsymbol{\hat{y}}_j = D_2([\boldsymbol{r}_{j,q}, \boldsymbol{r}_{j,2}])$. 
        \State Go to Round 1 and proceed with the next sample  $\boldsymbol{x}_{j+1}$.
    \Else
        \State Classify signals  $\boldsymbol{r}_{j,1}$, received over $n_{c,1}$ channel uses of Round 1 such that $\boldsymbol{\hat{y}}_j = D_1(\boldsymbol{r}_j)$. 
        \State Go to Round 1 and proceed with the next sample  $\boldsymbol{x}_{j+1}$.
    \EndIf
\end{algorithmic}
\end{algorithm}

The delay and the task accuracy depend on the threshold $\delta$. For a given threshold $\delta$, the delay for input sample $\boldsymbol{x}_j$ is given by 
\begin{equation}
    D_j = n_{c,1} \cdot \mathbbm{1} \left(\boldsymbol{y}_{j,1}^{\max} \geq \delta \right) + \left( n_{c,1} + n_{c,2} \right) \cdot \mathbbm{1} \left(\boldsymbol{y}_{j,1}^{\max} < \delta \right).
\end{equation}

Using $\boldsymbol{y}_1^{\max}$ for a given input sample. Then, the average delay is expressed as 
\begin{equation}
    \bar{D} = n_{c,1} \cdot P \left(\boldsymbol{y}_1^{\max} \geq \delta \right) + \left( n_{c,1} + n_{c,2} \right) \cdot P\left(\boldsymbol{y}_1^{\max} < \delta \right).
\end{equation}

For a given input sample, we denote $\boldsymbol{\hat{y}}_{m} $ as the random variable corresponding to the predicted label in round $m=1,2$. Then the average task accuracy (namely, the probability of successful classification) with MRMTL is given by 
\begin{eqnarray}
    P_S \hspace{-0.15cm} &=& \hspace{0.135cm} P\left( \boldsymbol{\hat{y}}_{1} = \boldsymbol{y}^* \: \vert \: \boldsymbol{y}_1^{\max} \geq \delta \right) \cdot P\left(  \boldsymbol{y}_1^{\max} \geq \delta \right)  \nonumber \\  & & \hspace{-0.2cm} + \:\:  P\left( \boldsymbol{\hat{y}}_{2} = \boldsymbol{y}^*  \: \vert \:  \boldsymbol{y}_1^{\max} < \delta \right) \cdot P\left(\boldsymbol{y}_1^{\max} < \delta \right).
\end{eqnarray}

\begin{figure}[t!]
\vspace{-0.4cm}
	\centering
	\includegraphics[width=0.925\columnwidth]{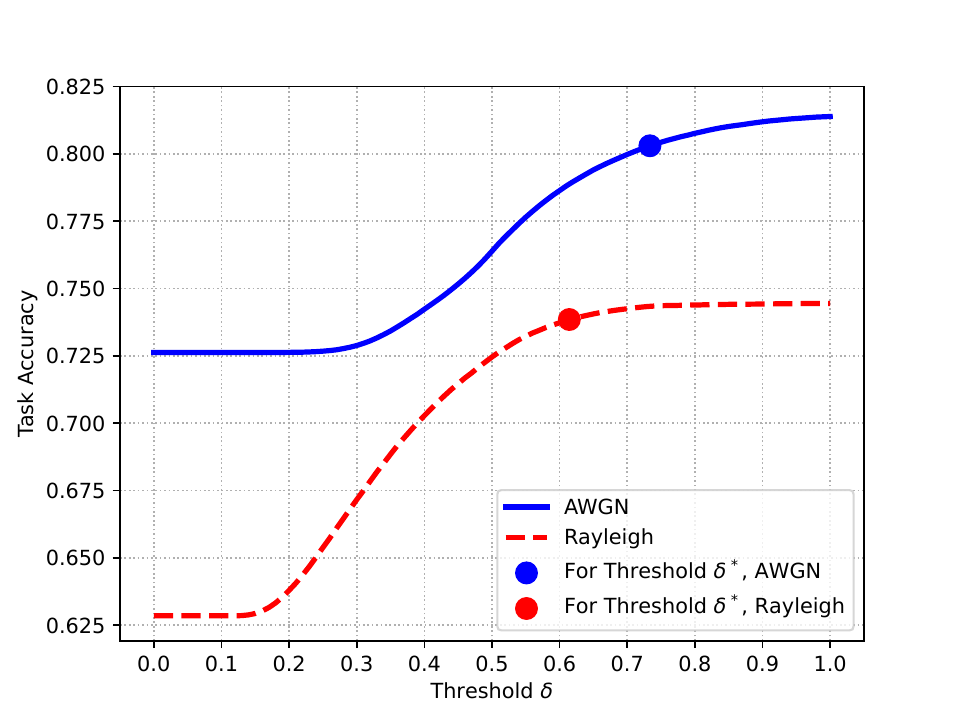}
    \caption{Task accuracy vs. threshold for round selection.}
	\label{fig:thr_succrate}
\end{figure}

\begin{figure}[ht]
\vspace{-0.5cm}
	\centering
	\includegraphics[width=0.925\columnwidth]{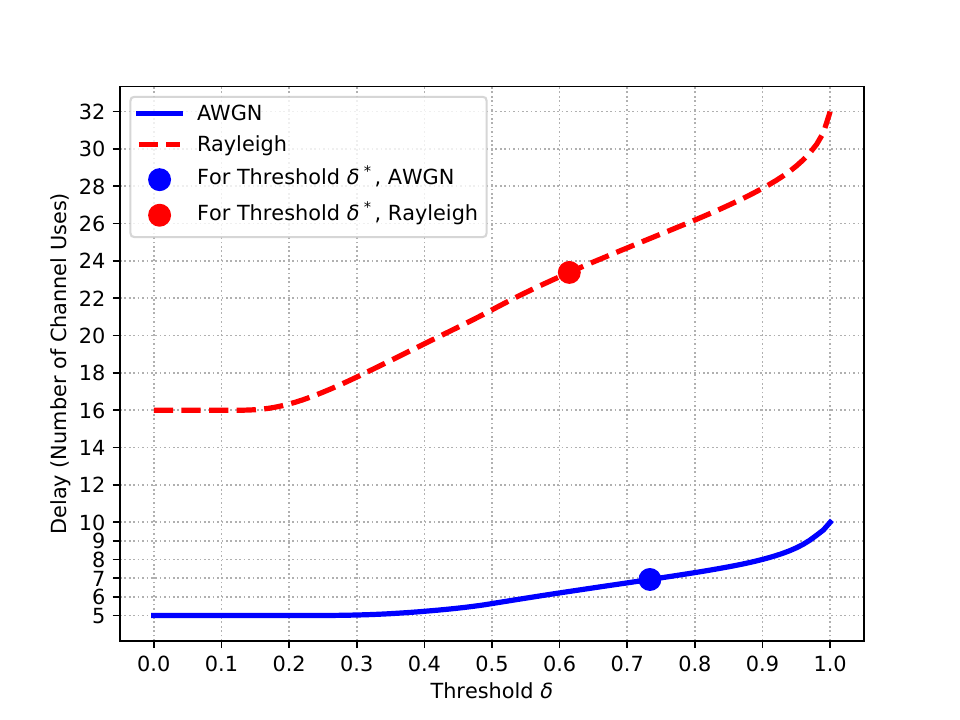}
    \caption{Delay (number of channel uses) vs. threshold for round selection.}
	\label{fig:thr_delay}
\end{figure}

\begin{figure}[ht]
\vspace{-0.5cm}
	\centering
    \includegraphics[width=0.925\columnwidth]{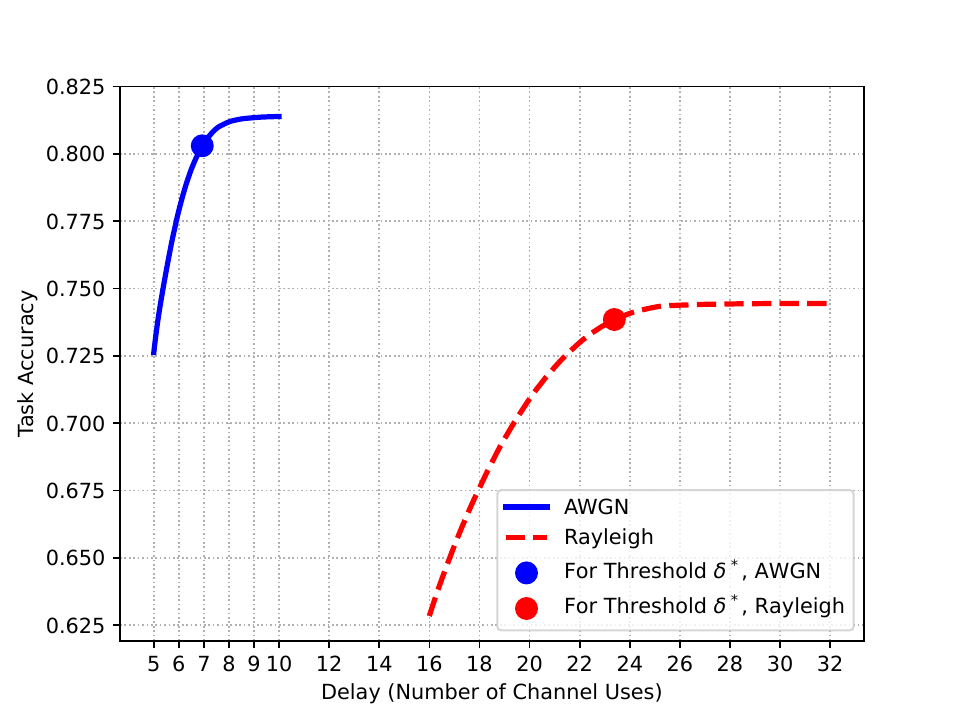}
    \caption{Task accuracy vs. delay (number of channel uses).}
	\label{fig:delay_succrate}
\end{figure}

Task accuracy and delay (total number of channel uses) are shown in Figs.~\ref{fig:thr_succrate} and ~\ref{fig:thr_delay} as a function of the threshold $\delta$ for round selection under AWGN and Rayleigh channels. Overall, task accuracy is higher and delay is lower when AWGN channel is used compared to Rayleigh channel. As the threshold $\delta$ increases, the likelihood of initiating Round 2 increases such that both the task accuracy and delay increase. Note that the task accuracy saturates as the threshold $\delta$ increases, indicating that a large threshold is not useful to further increase task accuracy but significantly increases the delay.
The underlying relationship between the task accuracy and delay is shown in Fig.~\ref{fig:delay_succrate}  under AWGN and Rayleigh channels. We observe operational modes, where the threshold $\delta$ can be gradually increased until a given target level of delay is reached. If further  feedback is available on the task accuracy performance achieved over time, the threshold $\delta$ can be also gradually increased until the target value of task accuracy is reached.

From Figs.~\ref{fig:histA} and \ref{fig:histR}, the average values of $\boldsymbol{y}_1^{\max}$  are shown in Table~\ref{tab:dec_process} for AWGN and Rayleigh channels, respectively. An effective separation can be obtained when threshold $\delta^*$ is selected as 
\begin{equation}
\delta^* = \frac{1}{2} \left( E[ \boldsymbol{y}_1^{\max} \: \vert \:  \boldsymbol{\hat{y}}_{1} = \boldsymbol{y}^*] + E[ \boldsymbol{y}_1^{\max} \: \vert \:  \boldsymbol{\hat{y}}_{1} \neq \boldsymbol{y}^*]   \right).
\end{equation}

Table~\ref{tab:dec_process} lists the threshold $\delta^*$ for multi-round selection that is also plotted in  Figs.~\ref{fig:histA} and \ref{fig:histR} for AWGN and Rayleigh channels, respectively. Table~\ref{tab:performance} highlights the high accuracy and low delay performance of MRMTL when the threshold $\delta$ is selected as $\delta^*$. Overall, task accuracy is higher under the AWGN channel compared to Rayleigh channel.
\label{tab:performance}
\begin{table}[ht]
\footnotesize
    \centering
    \caption{Threshold selection and performance of MRMTL.}
    \label{tab:dec_process_performance}
\begin{subtable}[t]{0.5\textwidth}
    \centering
    \caption{Threshold selection of MRMTL.}
    \label{tab:dec_process}
    \begin{tabular}{c||c|c}
     & AWGN & Rayleigh \\
     & channel & channel \\
    \hline
    Average of decoder final layer output  & & \\ for correct decisions, $E[y_c]$ & 0.8206 & 0.7489     \\   
    Average of decoder final layer output  & &  \\ for incorrect decisions, $E[y_i]$ & 0.6463 & 0.4798 \\ 
    Threshold for multi-round selection, $\delta^*$  & 0.7335 & 0.6143 \\ 
    \hline 
    \end{tabular}
    \vspace{0.5cm}
    \end{subtable}
    \vspace{0.5cm}
    \begin{subtable}[t]{0.5\textwidth}
    \centering
    \caption{Performance of MRMTL when threshold is selected as $\delta^*$.}
    \begin{tabular}{c||c|c}
    Performance & AWGN channel & Rayleigh channel \\
    \hline
    Accuracy  & 0.8030 & 0.7385 \\
    Delay & 6.9343 & 23.3842 \\ 
    \hline 
    \end{tabular}
    \end{subtable}
    
\end{table}

\section{Conclusion}
In this paper, we considered a TOC framework for NextG systems, where an encoder at the transmitter learns compressed latent representations of data and a decoder at the receiver performs classification tasks. By jointly training the encoder-decoder pair with DNNs, we optimized the transmission process to achieve high accuracy with minimal channel uses. Our novel approach, MRMTL, involves dynamically updating channel uses through iterative transmissions over multiple rounds, allowing the receiver to utilize signals from a previous round to enhance task performance. MRMTL, optimized through multi-task learning, balances accuracy and delay, and dynamically determines the need for additional channel uses by evaluating the confidence of the receiver in task decisions. Performance evaluation using the CIFAR-10 dataset and CNNs under AWGN and Rayleigh channel models demonstrates that MRMTL achieves the accuracy of conventional methods (that require more channel uses) but with significantly reduced delay. As a result, MRMTL significantly improves the efficiency of TOC, effectively balancing accuracy and delay. Low-latency design of MRMTL holds a strong potential in advancing TOC for NextG networks, making them more efficient and effective for applications such as IoT, AR/VR, and V2X networks. Future research may explore additional machine learning techniques and broader application domains to further enhance these strategies.

\end{document}